\documentclass[final]{cvpr}
\usepackage{times}
\usepackage{epsfig}
\usepackage{float}
\usepackage{subfigure}
\usepackage{graphicx}
\usepackage{amsmath}
\usepackage{amssymb}
\usepackage{bm}
\usepackage[pagebackref=true,breaklinks=true,colorlinks,bookmarks=false]{hyperref}

\def\cvprPaperID{2827} % *** Enter the CVPR Paper ID here
\def\confYear{CVPR 2021}
\title{One Shot Face Swapping on Megapixels}

\author{
Yuhao Zhu \textsuperscript{\rm 1}, \
Qi Li \thanks{Corresponding author} \space \textsuperscript{\rm 1,2}, \ %\footnotemark[1], \ 
Jian Wang \textsuperscript{\rm 1,3},  \
Chengzhong Xu \textsuperscript{\rm 2}, \
Zhenan Sun\textsuperscript{\rm 1,3}\\
\textsuperscript{\rm 1}Center for Research on Intelligent Perception and Computing, NLPR, CASIA\\
\textsuperscript{\rm 2}State Key Laboratory of IoTSC,  Faculty of Science and Technology, University of Macau\\
\textsuperscript{\rm 3}School of Artificial Intelligence, University of Chinese Academy of Sciences\\
\tt\small \{yuhao.zhu, jian.wang\}@cripac.ia.ac.cn, \{qli, znsun\}@nlpr.ia.ac.cn, czxu@um.edu.mo
}
\begin{document}
%\footnote[1]{Corresponding author.}
\twocolumn[{
\maketitle
\thispagestyle{empty}
\small
\begin{figure}[H]
    \vspace{-1cm}
    \hsize=1.0\textwidth
    \begin{center}
    \includegraphics[width=2.1\linewidth]{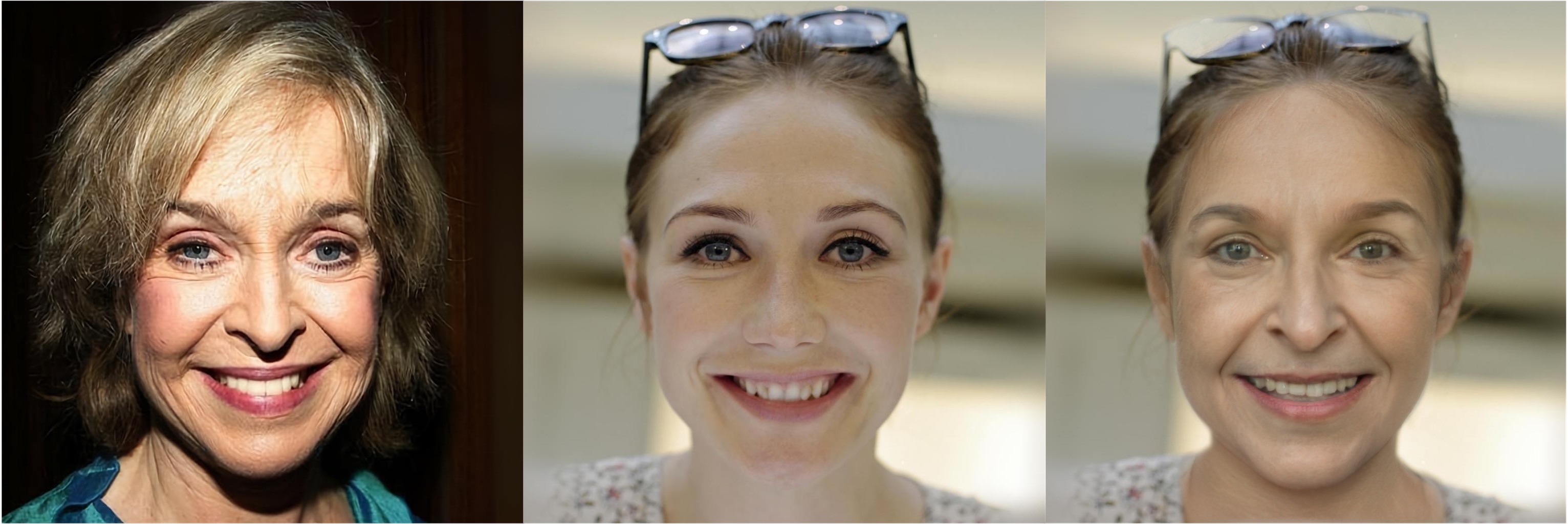}
    \caption{Example of a swapped face. Left: source image that represents the identity; Middle: target image that provides the attributes; Right: the swapped face image. All images are in $1024^2$.}
    \label{fig:snapshot}
    \end{center}
    \end{figure}
}]
{
    \renewcommand{\thefootnote}{\fnsymbol{footnote}}
    \footnotetext[1]{Corresponding author}
}

%\thispagestyle{empty}

%%%%%%%%% ABSTRACT
\begin{abstract}
    \vspace{-0.355cm}
    Face swapping has both positive applications such as entertainment, human-computer interaction, etc., and negative applications such as DeepFake threats to politics, economics, etc. Nevertheless, it is necessary to understand the scheme of advanced methods for high-quality face swapping and generate enough and representative face swapping images to train DeepFake detection algorithms. This paper proposes the first Megapixel level method for one shot Face Swapping (or MegaFS for short). Firstly, MegaFS organizes face representation hierarchically by the proposed Hierarchical Representation Face Encoder (HieRFE) in an extended latent space to maintain more facial details, rather than compressed representation in previous face swapping methods. Secondly, a carefully designed Face Transfer Module (FTM) is proposed to transfer the identity from a source image to the target by a non-linear trajectory without explicit feature disentanglement. Finally, the swapped faces can be synthesized by StyleGAN2 with the benefits of its training stability and powerful generative capability. Each part of MegaFS can be trained separately so the requirement of our model for GPU memory can be satisfied for megapixel face swapping. In summary, complete face representation, stable training, and limited memory usage are the three novel contributions to the success of our method. Extensive experiments demonstrate the superiority of MegaFS and the first megapixel level face swapping database is released for research on DeepFake detection and face image editing in the public domain. The dataset is at this \href{https://github.com/zyainfal/One-Shot-Face-Swapping-on-Megapixels}{link}.
\end{abstract}
\vspace{-0.5cm}
\section{Introduction}
Given two face images, face swapping refers to transferring the identity from the source image to the target image, while the facial attributes of the target image hold intact. It has attracted extensive attention in recent years for its broad application prospects in entertainment \cite{alexander2009creating, kemelmacher2016transfiguring}, privacy protection \cite{blanz2004exchanging, mosaddegh2014photorealistic}, and theatrical industry \cite{naruniec2020high}.

Existing face swapping methods can be roughly divided into two categories: subject-specific and subject agnostic methods. Subject-specific face swapping methods \cite{Ondyari2019DeepFakes, korshunova2017fast, naruniec2020high} need to be trained and tested on the same pair of subjects, which restricts their potential applications. On the contrary, subject agnostic face swapping methods \cite{natsume2018fsnet, bao2018towards, li2019faceshifter, nitzan2020facedisentangling, nirkin2019fsgan} can be applied to arbitrary identities without additional training procedures. In this paper, we focus on a more challenging topic: \emph{one shot face swapping}, where only one image is given from the source and target identity for both training and testing.

With the rapid growth of high resolution image and video data on the web, it becomes increasingly popular to process high resolution samples. However, generating high resolution swapped faces is rather difficult because of the following problems. Firstly, information is insufficient for high-quality face generation due to the compressed representation in an end-to-end framework \cite{bao2018towards, natsume2018fsnet, li2019faceshifter}. Secondly, adversarial training is unstable \cite{chen2017photographic}, which confines the resolution of previous methods only up to $256^2$. Thirdly, the GPU memory limitation makes the training untenable, or the training batch is bounded by a small size, which aggravates the collapse of the training process.
%Generally, a unified framework with complex pre-processes and post-processes dominates previous subject agnostic face swapping methods. However, the end-to-end design of these works fails to generate high resolution swapped faces due to unstable adversarial training and GPU memory limitations.

%Face swapping methods usually consist of an encoder, a swapping module and a decoder. In this paper, we develop a novel three-stage learning framework for high resolution face swapping by adopting a ``divide and conquer'' strategy.
To this end, this paper proposes the first Megapixel level one shot Face Swapping method (MegaFS) by adopting the ``divide and conquer'' strategy in three steps.
Firstly, to overcome the information loss in the encoder, we adopt GAN Inversion methods \cite{perarnau2016invertible, creswell2018inverting, guan2020collaborative, abdal2019image2stylegan, abdal2020image2stylegan++, creswell2018inverting, lipton2017precise, Ondyari2019DeepFakes, zhu2020domain} and propose a Hierarchical Representation Face Encoder (HieRFE) to find the complete face representation in an extended latent space $\mathcal{W}^{++}$. Secondly, to modify face representations and resolve the problem of previous latent code manipulation methods \cite{radford2015unsupervised, harkonen2020ganspace, shen2020interpreting, tewari2020stylerig, tewari2020pie, blanz1999morphable, nitzan2020disentangling, abdal2020styleflow} that only one attribute can be modified once a time, a novel swapping module, Face Transfer Module (FTM), is proposed to control multiple attributes synchronously without explicit feature disentanglement. Finally, the unstable adversarial training problem is evaded by exploiting StyleGAN2 \cite{karras2020analyzing} as the decoder, which is fixed and the discriminator is not used for optimization. Each part of MegaFS can be trained separately so the GPU memory requirements are satisfied for megapixel face swapping. The contributions of this paper can be summarized as:

\begin{itemize}
    \item To the best of our knowledge, the proposed MegaFS is the first method that can conduct one shot face swappings at megapixel level.
\end{itemize}

\begin{itemize}
    \item For encoding and manipulating the complete face representation, faces are encoded by HieRFE hierarchically in the new extended latent space $\mathcal{W}^{++}$ and a new multistep non-linear latent code manipulation module, FTM, is proposed to manage multiple attributes synchronously without explicit feature disentanglement.
    %The problems of unstable adversarial training and GPU memory limitation are resovled by exploiting StyleGAN2 and separated training procedure.
    %We propose a hierarchical representation for complete face representation in a new latent space $\mathcal{W}^{++}$, which shows better stability and controllability for face swapping. Besides, to satisfy the requirements of face swapping, a new multistep non-linear latent code manipulation module, Face Transfer Module (FTM), is proposed to manage multiple attributes synchronously.
\end{itemize}

\begin{itemize}
    \item Experimental results on benchmark dataset have shown the effectiveness of the proposed MegaFS. Furthermore, the first megapixel face swapping database is released for research of DeepFake detection and face image editing in the public domain.
    %Experimental results on the benchmark dataset have shown the superiority of our method. It performs competitively when compared with other state-of-the-art face swapping methods at $256^2$. Furthermore, we have built a face swapping dataset based on CelebA-HQ and released a baseline for megapixel level face swapping.
\end{itemize}

%In the rest of this paper, we will introduce the related works of GAN based researches and face swapping advances in section 2. The details of the proposed pipeline are in section 3.
%Experiments and results are presented in section 4. Finally, we conclude in section 5.

\section{Related Works}
%Since Generative Adversarial Networks introduced by Goodfellow \etal \cite{goodfellow2014generative}, a huge amount of follow up works sparked. Current state-of-the-art GANs are proposed for attribute editing \cite{he2019attgan, choi2018stargan, choi2020stargan}, Image-to-Image \cite{CycleGAN2017, pix2pix2017}, and high fidelity image generation \cite{karras2017progressive, karras2019style, karras2020analyzing}.
%Based on a well trained GAN model, the pioneers of GAN Inversion \cite{creswell2018inverting} and  Latent Space Manipulation \cite{shen2020interpreting} studied the latent space of \cite{karras2017progressive, karras2019style, karras2020analyzing} for better GAN understanding and controlling.
%In the meanwhile, face swapping demonstrated it irresistible attractions, and GAN based frameworks have been well studied at the resolution of $256^2$.
\subsection{Face Swapping}
Subject-specific face swapping methods are popular in recent years, where DeepFake \cite{Ondyari2019DeepFakes} and its variants are trained using pairwise samples. Besides, Korshunova \etal \cite{korshunova2017fast} model different source identities separately, such as a CageNet for Nicolas Cage, or a SwiftNet for Taylor Swift.
Recently, Disney Research realizes high resolution face swapping \cite{naruniec2020high}, but it requires training decoders for different subjects, which hinders its generalization. Besides, it is time consuming and difficult for subject-specific methods to train specific models for distinct pairs of faces \cite{dolhansky2019deepfake, roessler2019faceforensicspp, rossler2018faceforensics, Celeb_DF_cvpr20, jiang2020deeperforensics, yang2019exposing, korshunov2018deepfakes}.
Subsequently, subject agnostic face swapping methods break the limitations of previous subject-specific face swapping methods. Realistic Neural Talking Head \cite{olszewski2017realistic} adopts meta-learning to relieve the pain of fine-tuning on different individuals. FaceSwapNet \cite{zhang2019faceswapnet} proposes a landmark swapper to handle the identity leakage problem from landmarks. 
%Recently, FaR GAN \cite{hao2020far} borrows ideas from style transfer \cite{huang2017arbitrary, park2019semantic} to inject identity information into features from convoluted landmarks. 
In the meanwhile, other mindsets follow the attribute disentanglement heuristic to explore new high fidelity face swapping frameworks. FSNet \cite{natsume2018fsnet} represents the face region of the source image as a vector, which is combined with a non-face target image to generate the swapped face image. IPGAN \cite{bao2018towards} disentangles identities and facial attributes as different vectorized representaions. Based on previous works, FSGAN \cite{nirkin2019fsgan} and FaceShifter \cite{li2019faceshifter} achieve state-of-the-art results by their outstanding performance.

\subsection{GAN Inversion}
Based on a well-trained GAN, GAN Inversion, or Latent Space Embedding, tries to find the latent code that can accurately reconstruct a given image synthesized. To this end, two problems need to be settled: determining a proper latent space and designing an algorithm to search for the optimal latent code within that space. As for the latent space, early methods perform image inversion into $\mathcal{W} \in \mathbb{R}^{1 \times 512}$ \cite{perarnau2016invertible, jahanian2019steerability, harkonen2020ganspace}, while later works \cite{abdal2019image2stylegan, abdal2020image2stylegan++, Ondyari2019DeepFakes, creswell2018inverting} extend the latent space to $\mathcal{W}^+ \in \mathbb{R}^{18 \times 512}$, which proves to have better reconstruction results.
As for the inversion algorithms, they either train an encoder \cite{perarnau2016invertible, creswell2018inverting, guan2020collaborative} to predict latent codes of images or minimize the error between predicted and given images by optimizing latent codes from random initializations \cite{abdal2019image2stylegan, abdal2020image2stylegan++, creswell2018inverting, lipton2017precise}. Some methods \cite{Ondyari2019DeepFakes, zhu2020domain} combine both to optimize latent codes initialized by encoders.

%According to the statement in \cite{richardson2020encoding}, the lower level latent codes in $\mathcal{W}^+$ are mainly responsible for the topology control in a generated face image. However, the lowest latent code for semantic representations, the $4 \times 4 \times 512$ \emph{constant} input of the StyleGAN2 \cite{karras2020analyzing}, remains intact. We will show the \emph{constant} is also controllable for better reconstruction quality. Therefore, the latent space is extended to $\mathcal{W}^{++}$ by introducing the \emph {constant} in this paper.

\begin{figure*}[t]
    \vspace{-0.7cm}
    \begin{center}
    \includegraphics[width=1.0\linewidth]{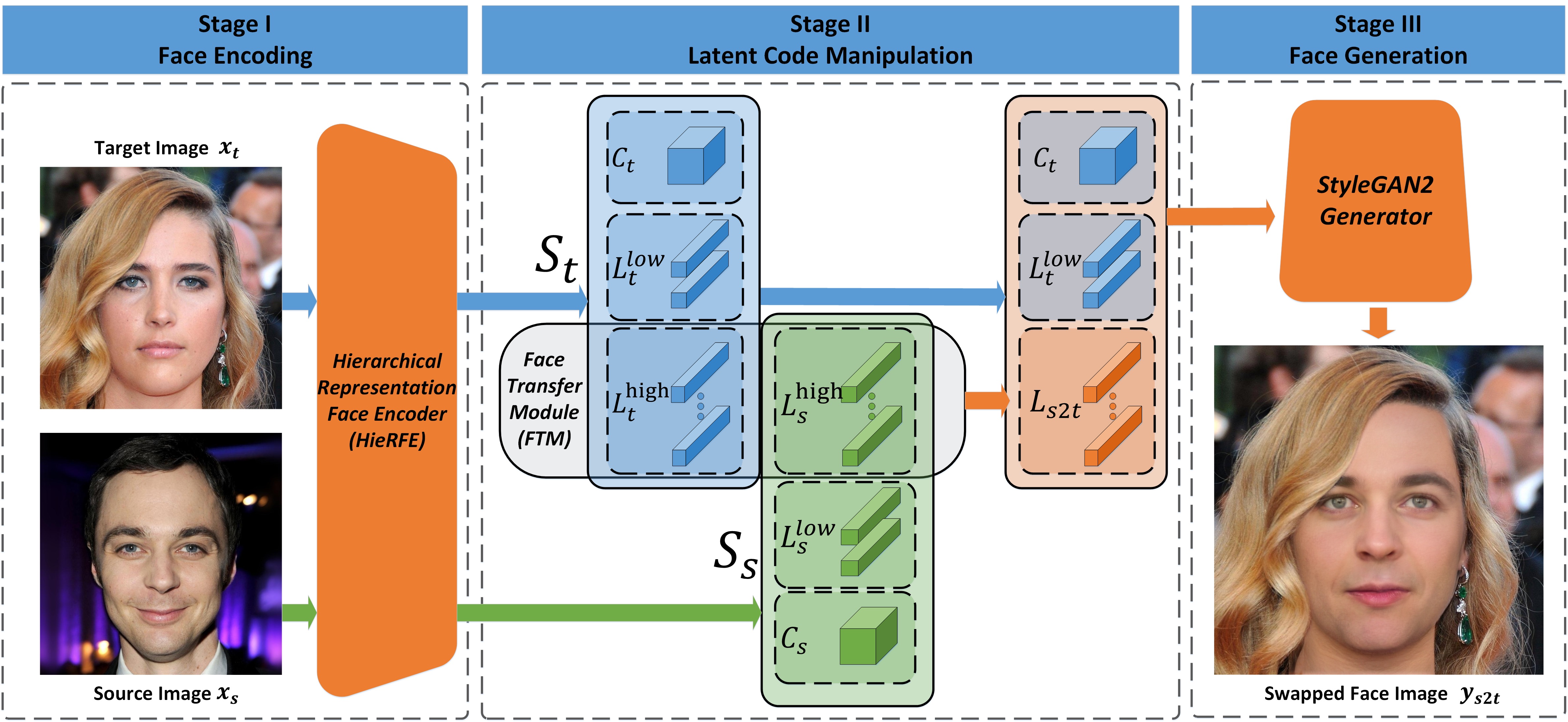}
    \caption{The proposed MegaFS consists of three stages: Face Encoding, Latent Code Manipulation, and Face Generation. Firstly, HieRFE projects two face images into latent space $\mathcal{W}^{++}$. Then FTM manipulates $L_s^{high}$ and $L_t^{high}$ in two hierarchical latent sets $S_s$ and $S_t$ to get $L_{s2t}$. Finally, the swapped face image $y_{s2t}$ can be synthesized by a pre-trained StyleGAN2 generator from $C_t$, $L_t^{low}$, and $L_{s2t}$.}
    \label{fig:pipeline}
\end{center}
\end{figure*}
\begin{figure}[]
    \vspace{-0.2cm}
    \begin{center}
    \includegraphics[width=1.0\linewidth]{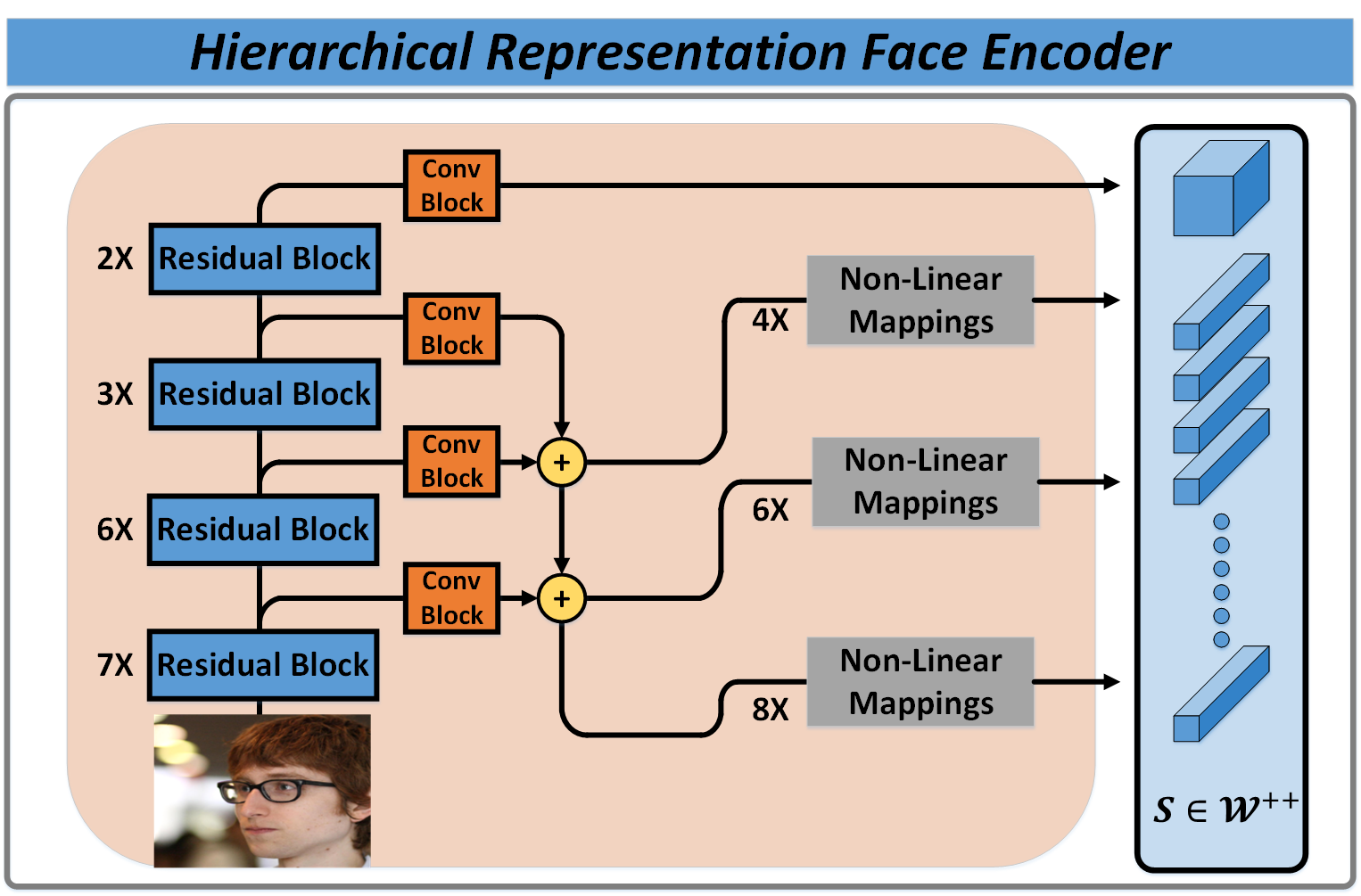}
    \caption{HieRFE consists of a ResNet50 backbone based on residual blocks, a feature pyramid structure based on FPN, and eighteen lateral non-linear mapping networks, in which $n \times$ refers to the number of the corresponding parts.}
    \label{fig:gie}
    \end{center}
\end{figure}

\subsection{Latent Code Manipulation}
Latent Code Manipulation, or Latent Control, is another attractive research area to manipulate latent codes based on the observation that semantic editing operations can be realized by adding high dimensional directions \cite{radford2015unsupervised}. Several linear semantic directions, or trajectories, of $\mathcal{W}$ are found \cite{harkonen2020ganspace, shen2020interpreting}. StyleRig \cite{tewari2020stylerig} and PIE \cite{tewari2020pie} propose to manipulate latent space through an existing 3D model \cite{blanz1999morphable}, which successfully control facial poses, expressions, and illuminations. Previous methods \cite{harkonen2020ganspace, tewari2020stylerig, nitzan2020disentangling, radford2015unsupervised} have found good controllability of StyleGAN based on the assumption that semantic directions in StyleGAN latent space are linear. Recently, StyleFlow \cite{abdal2020styleflow} achieves better manipulation results through non-linear trajectories.%, which have shown the potential of latent control in an non-linear space.

%While all methods above are able to change the target attributes in a generated image, it is not feasible to adopt them directly to the face swapping task as they can only change one attribute once a time. On the contrary, in this paper, we propose an ingenious design module to search non-linear trajectories to control multiple attributes synchronously, as required by the face swapping task.

\section{Method} \label{sec:method}
%\subsection{Megapixel Level Face Swapping} \label{subsec:megafs}
Fig.\ref{fig:pipeline} demonstrates the overall pipeline and notations of the proposed MegaFS, which combines the identity information from a source image $x_s$ and attribute information from a target image $x_t$ to generate the final swapped face image $y_{s2t}$. In the following, we will present the details of our method.

\subsection{Hierarchical Representation for Face Swapping}
In the first stage, face images are projected into latent space $\mathcal{W}^{++}$ using Hierarchical Representation Face Encoder (HieRFE) to deposit complete face information. 
The structure of HieRFE is detailed in Fig.\ref{fig:gie}.

Specifically, HieRFE consists of a ResNet50 backbone based on several residual blocks \cite{he2016deep}, a feature pyramid structure based on FPN \cite{lin2017feature} for feature refinement, and eighteen lateral non-linear mapping networks for latent code prediction. Please refer to the corresponding papers for details of residual blocks and FPN. As for the non-linear mapping network, it comprises repeated downsampling, convolution, batchnorm, and leakyReLU layers until the feature map can be pooled as a vector, \ie, $l \in \mathbb{R}^{1 \times 512}$. 

Then, the constant input of StyleGAN2 predicted by the backbone and four latent codes predicted by the smallest feature map, denoted as $C \in \mathbb{R}^{4 \times 4 \times 512}$ and $L^{low} \in \mathbb{R}^{4 \times 512}$, represent low-level topology information. Other latent codes are gathered as $L^{high} \in \mathbb{R}^{14 \times 512}$ to represent high-level semantic information. Finally, subscript $s$ and $t$ are adopted to represent the source and target images if it is necessary in the following paper.

\subsection{Synchronized Control of Multiple Attributes}
During the second stage, Face Transfer Module (FTM) is proposed to control multiple attributes of identity information in a synchronized manner for face swapping demands. 
In detail, FTM contains $14$ Face Transfer Blocks, the number of which equals that of $l^{high}$. 
%Specifically, identity information is transferred from $x_s$ to $x_t$ by feeding $L_s^{high}$ and $L_t^{high}$ to the proposed Face Transfer Modules (FTM) to get the transferred high-level semantic codes $L_{s2t} \in \mathbb{R}^{14 \times 512}$. 

\begin{figure}[]
    \begin{center}
    \includegraphics[width=1.0\linewidth]{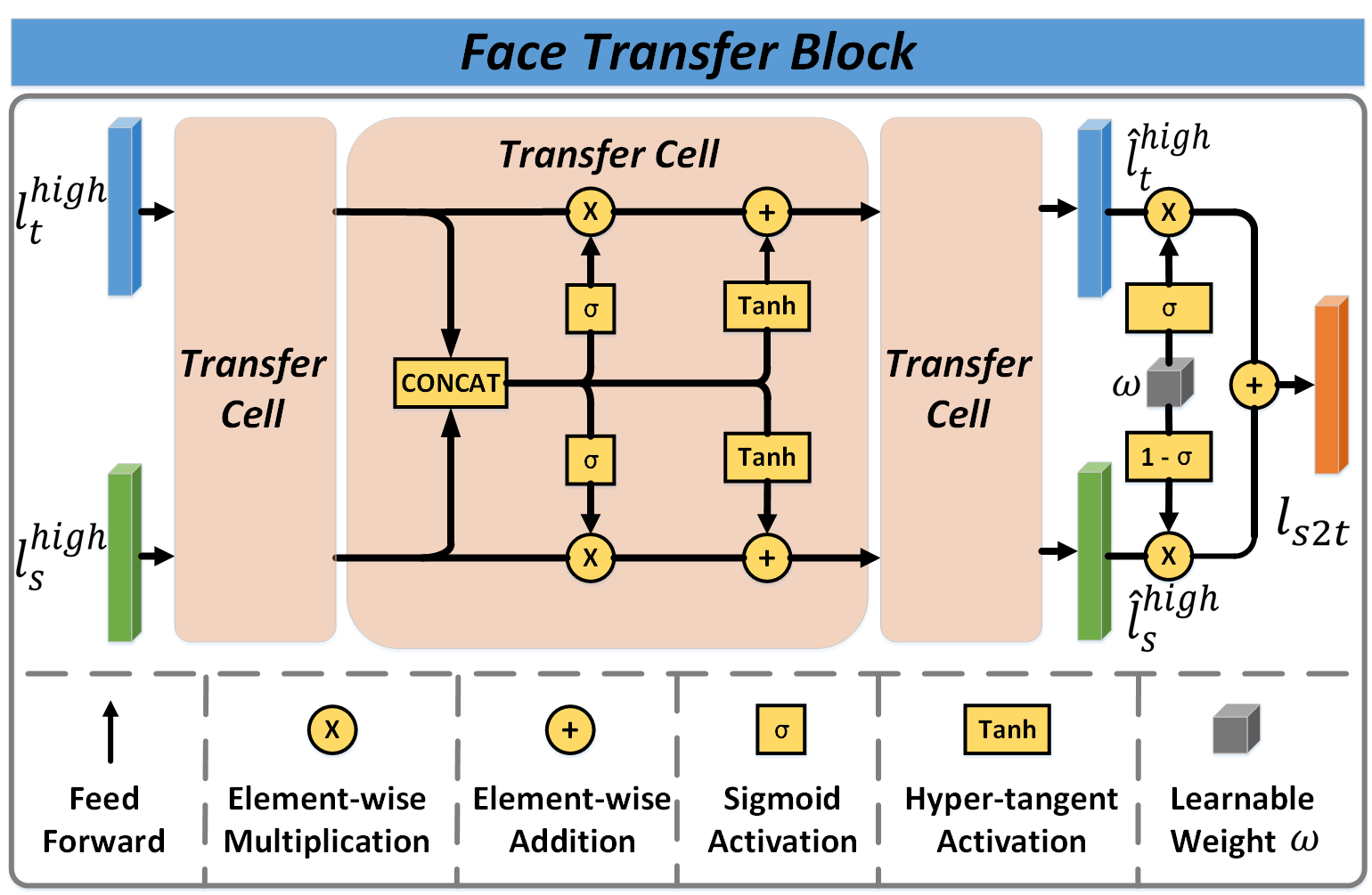}
    \caption{Inside FTM, each Face Transfer Block contains three identical transfer cell. After being processed by three cells, two refined vectors are weighted by a learnable weight $\omega$ and summed as the final output.}
    \label{fig:ftm}
    \end{center}
\end{figure}

As shown in Fig.\ref{fig:ftm}, each Face Transfer Block contains three identical transfer cells. In each transfer cell, $l_s^{high}$ and $l_t^{high}$ are firstly concatenated to $l_c^{high}$, which collects all information from the source and target images. Then $l_s^{high}$ is refined to $\hat{l}_s^{high}$ through a two-step non-linear trajectory:
\begin{equation}
    \mathcal{T}(l_c^{high}, l_s^{high}) = \mathcal{T}_2(l_c^{high}, \mathcal{T}_1(l_c^{high}, l_s^{high}))
\end{equation}
in which
\begin{flalign}
    \begin{aligned}
        &\mathcal{T}_1(a, b) = sigmoid(K_1(a)) \times b \\
        &\mathcal{T}_2(a, b) = Tanh(K_2(a)) + b
    \end{aligned}
\end{flalign}
where $K_1(\cdot)$ and $K_2(\cdot)$ denote two linear layers. The trajectory is crafted based on the following illustrations. In the first step, the multiplication coefficients are scaled in range $(0, 1)$ after sigmoid activation, where $l_s^{high}$ is designed to discard irrelevant semantics except for the identity information. In the second step, $l_s^{high}$ accepts a small amount of target semantic attributes by shifting in the latent space.
Similarly, $l_t^{high}$ is processed in parallel but for discarding target identity while holding other semantics.
Finally, the transferred latent code $l_{s2t} \in L_{s2t}$ can be predicted as
\begin{equation}
    l_{s2t} = \sigma(\omega)\hat{l}_t^{high} + (1-\sigma(\omega))\hat{l}_s^{high}
\end{equation}
where $\omega \in \mathbb{R}^{1 \times 512}$ is a trainable weight vector, and $\sigma$ stands for the sigmoid activation. The transferred latent codes $L_{s2t}$ is composed by gathering all predicted $l_{s2t}$.

%Specifically, $L_s^{high}$ and $L_t^{high}$ are fed into FTM to generate $L_{s2t}$.
%in which fourteen $l_s^{high}$ and $l_t^{high}$ are diveded from $L_s^{high}$ and $L_t^{high}$ and processed by Face Transfer Blocks.
%At first, FTM splits $L_s^{high}$ and $L_t^{high}$ into fourteen $l_s^{high}$ and $l_t^{high}$ respectively.
%Afterwards, fourteen Face Transfer Blocks inside FTM process $l_s^{high}$ and $l_t^{high}$ in pairs.
%Specifically, $L_s^{high}$ and $L_t^{high}$ are splited into fourteen $l_s^{high}$ and $l_t^{high}$. Thus, fourteen Face Transfer Block are embedded in FTM to process pairwise $l_s^{high}$ and $l_t^{high}$.
\subsection{High-Fidelity Face Generation}
Finally in the third stage, $C_s$ and $L_s^{low}$ are discarded since they contain negligible identity information from $x_s$. The swapped face image $y_{s2t}$ can be generated by feeding StyleGAN2 generator with $C_t$, $L_t^{low}$ and $L_{s2t}$.

By taking StyleGAN2 as the decoder, face swapping through latent space differentiates our method from other face swapping frameworks. Firstly, it provides an extended latent space for complete face representation, which makes detailed face generation feasible. Secondly, it makes our method operating globally in $\mathcal{W}^{++}$ instead of locally on feature maps, which is desirable as it can conduct non-linear transformations through latent code manipulations without local distortions. Thirdly, it does not require explicit attributes disentanglement, which makes the training process straightforward without tricky loss functions and hyper-parameter settings.

%Moreover, previous works \cite{collins2020editing, karras2019style, karras2020analyzing} have demonstrated that the disentanglement of semantic objects learned by StyleGAN2 is due to its layer-wise representation, which makes face swapping feasible to operate only on $L^{high}$.
%Besides, a style may hide in multiple latent codes, which makes it intractable for users to modify a group of semantic representations synchronously and properly, especially for face swapping. Thus, FTM is proposed to transfer identity-related attributes from the source face to the target face in one inference process.
%By combining all the components above, we benefit from the powerful generative capability of StyleGAN2 over variants of facial attributes. Finally, the high-fidelity photorealistic swapped face images can be synthesized.
%from global representations of $\mathcal{W}^{++}$ and conquer the semantic manipulation difficulties for the face swapping task. Finally, high-fidelity photorealistic swapped face images can be synthesized by StyleGAN2.

\subsection{Objective Functions}
For each part of MegaFS, HieRFE and FTM are trained sequentially, while StyleGAN2 generator remains intact.

\vspace{0.2cm}
\noindent\textbf{Objective function of HieRFE:}
Following the previous work \cite{richardson2020encoding}, we make use of three objectives for supervising a pair of input image $x$ and its reconstruction image $\hat{x}$, including pixel-wise reconstruction loss $\mathcal{L}_{rec}$, Learned Perceptual Image Path Similarity (LPISP) loss $\mathcal{L}_{LPIPS}$ \cite{zhang2018unreasonable}, and identity loss $\mathcal{L}_{id}$ as follows:
\begin{equation}
    \mathcal{L}_{rec}   = \left \| x - \hat{x} \right \|_2
\end{equation}
\begin{equation}
    \mathcal{L}_{LPIPS} = \left \| F(x) - F(\hat{x}) \right \|_2
\end{equation}
\begin{equation}
    \mathcal{L}_{id}    = 1 - \cos(R(x), R(\hat{x}))
\end{equation}
\noindent where $\left \| \cdot \right \|_2$ denotes $\ell _2$ distance, $F(\cdot)$ denotes the perceptual feature extractor, $R(\cdot)$ denotes the ArcFace \cite{deng2019arcface} recognition model, $\cos (\cdot, \cdot)$ denotes the cosine similarity of two face embeddings. 

In addition, as face swapping needs pose and expression controllability, we introduce landmarks loss $\mathcal{L}_{ldm}$ to measure $\ell _2$ difference between the predicted landmarks of the input faces and the corresponding ones of reconstructed faces as following:
\begin{equation}
    \mathcal{L}_{ldm}   = \left \| P(x) - P(\hat{x}) \right \|_2
\end{equation}
where $P(\cdot)$ denotes the facial landmark predictor \cite{WangSCJDZLMTWLX19}.
The overall loss function for training HieRFE is
\begin{equation}
\mathcal{L}_{inv} = \lambda_{1}\mathcal{L}_{rec}
                + \lambda_{2}\mathcal{L}_{LPIPS}
                + \lambda_{3}\mathcal{L}_{id}
                + \lambda_{4}\mathcal{L}_{ldm}
\end{equation}

\noindent where $\lambda_{1}, \lambda_{2}, \lambda_{3}$ and $\lambda_{4}$ are loss weights. Besides, $x$ and $\hat{x}$ need resizing as the input of each model before calculating the loss function.

\vspace{0.2cm}
\noindent\textbf{Objective function of FTM:}
For training FTM, four losses are proposed, including:
\begin{equation}
    \mathcal{L}'_{rec} = \left \| x_s - \hat{x}_s \right \|_2 + \left \| x_t - \hat{x}_t \right \|_2
\end{equation}
\begin{equation}
    \mathcal{L}'_{LPIPS} = \left \| F(x_t) - F(y_{s2t}) \right \|_2
\end{equation}
\begin{equation}
    \mathcal{L}'_{id} = 1 - \cos(R(x_s), R(y_{s2t}))
\end{equation}
\begin{equation}
    \mathcal{L}'_{ldm} = \left \| P(x_t) - P(y_{s2t}) \right \|_2
\end{equation}
\noindent Besides, $\mathcal{L}_{norm}$ is leveraged to stabilizes the training process.
\begin{equation}
    \mathcal{L}_{norm} = \left \| L_s^{high} - L_{s2t} \right \|_2
\end{equation}
\noindent Similarly, the overall loss function for training FTM is
%\begin{flalign}
%    \begin{aligned}
    \begin{equation}
        \mathcal{L}_{swap} = \varphi_{1}\mathcal{L}'_{rec}
                    + \varphi_{2}\mathcal{L}'_{LPIPS}
                    + \varphi_{3}\mathcal{L}'_{id}
                    + \varphi_{4}\mathcal{L}'_{ldm}
                    + \varphi_{5}\mathcal{L}_{norm}
                \end{equation}
%    \end{aligned}
%\end{flalign}
\noindent where $\varphi_{1}, \varphi_{2}, \varphi_{3}, \varphi_{4}$ and $\varphi_{5}$ are loss weights. Finally, when FTM converges, the proposed method is ready for face swapping on megapixels.
%%%%%%%%% BODY TEXT

%-------------------------------------------------------------------------
\section{Experiments} \label{sec:exp}
In this section, we will first show the effectiveness of the proposed method by comparing it with other state-of-the-art methods provided in FaceForensics++ \cite{roessler2019faceforensicspp}. Then the superiority of our method is demonstrated by conducting face swapping on CelabA-HQ \cite{karras2017progressive}. Finally, an ablation study is presented to reveal the necessity of each component of our method.

\subsection{Datasets and Implementation Details} \label{subsec:did}
\noindent \textbf{CelebA \cite{liu2015faceattributes}: }This dataset is built for face detection, facial landmark localization, attribute recognition and control, and face synthesis. It contains 202,599 celebrity images with 40 labeled attributes and 5 landmark location annotations.

\noindent \textbf{CelebA-HQ \cite{karras2017progressive}: }It is a high-quality version of CelebA dataset. All 202,599 images in CelebA are processed by two pre-trained neural nets for denoising and super-resolution, resulting in 30,000 high-quality images.

\noindent \textbf{FFHQ \cite{karras2019style}: }The dataset contains 70,000 megapixel face images collected from Flickr. FFHQ has considerable variations of age, ethnicity, gender, and background. %It is also the dataset used to train StyleGAN2.

\noindent \textbf{FaceForensics++ \cite{roessler2019faceforensicspp}: }It is a forensics dataset consisting of 1,000 original video sequences from YouTube that have been manipulated with five automated face manipulation methods: Deepfakes, Face2Face, FaceSwap, NeuralTextures, and FaceShifter, in which Deepfakes, FaceSwap, and FaceShifter are face swapping methods, while Face2Face and NeuralTextures are reenactment algorithms.

\noindent \textbf{Implementation Details:} In all experiments, learning rate of the Adam optimizer \cite{kingma2014adam} is set to 0.01. We set $\lambda_{1}, \lambda_{2}, \lambda_{3}$ and $\lambda_{4}$ to 1, 0.8, 1, and 1000. We set $\varphi_{1}, \varphi_{2}, \varphi_{3}, \varphi_{4}$ and $\varphi_{5}$ to 8, 32, 24, 100000, and 32. In addition, 200,000 faces are randomly sampled as auxiliary data by running StyleGAN2.

For experiments on FaceForensics++, HieRFE and FTM are sequentially trained ten epochs in total on CelebA, CelebA-HQ, FFHQ, and the auxiliary data. As for experiments on CelebA-HQ, HieRFE and FTM are sequentially trained seventeen epochs in total on FFHQ and the auxiliary data. As for training time, it takes about five days on three Tesla V100 GPUs.
%For experiments on FaceForensics++, training set contains CelebA, CelebA-HQ, FFHQ, and the auxiliary data. The model is trained five epochs for both GAN Inversion and face swapping tasks.
%As for the experiments on CelebA-HQ, training set contains FFHQ and the auxiliary data. The model is trained five epochs and twelve epochs for the GAN Inversion and face swapping tasks respectively. Training MegaFS at $1024^2$ on three Tesla V100 GPUs takes about five days.

\subsection{Experiments on FaceForensics++} \label{subsec:expffpp}
\noindent \textbf{Qualitative Comparison:}
As FaceForensics++ contains images generated by three face swapping methods: FaceSwap, DeepFakes, and FaceShifter, we extract frames of the same index from this dataset and compare them with the proposed MegaFS.
\begin{figure}[]
    \begin{center}
    \includegraphics[width=1.0\linewidth]{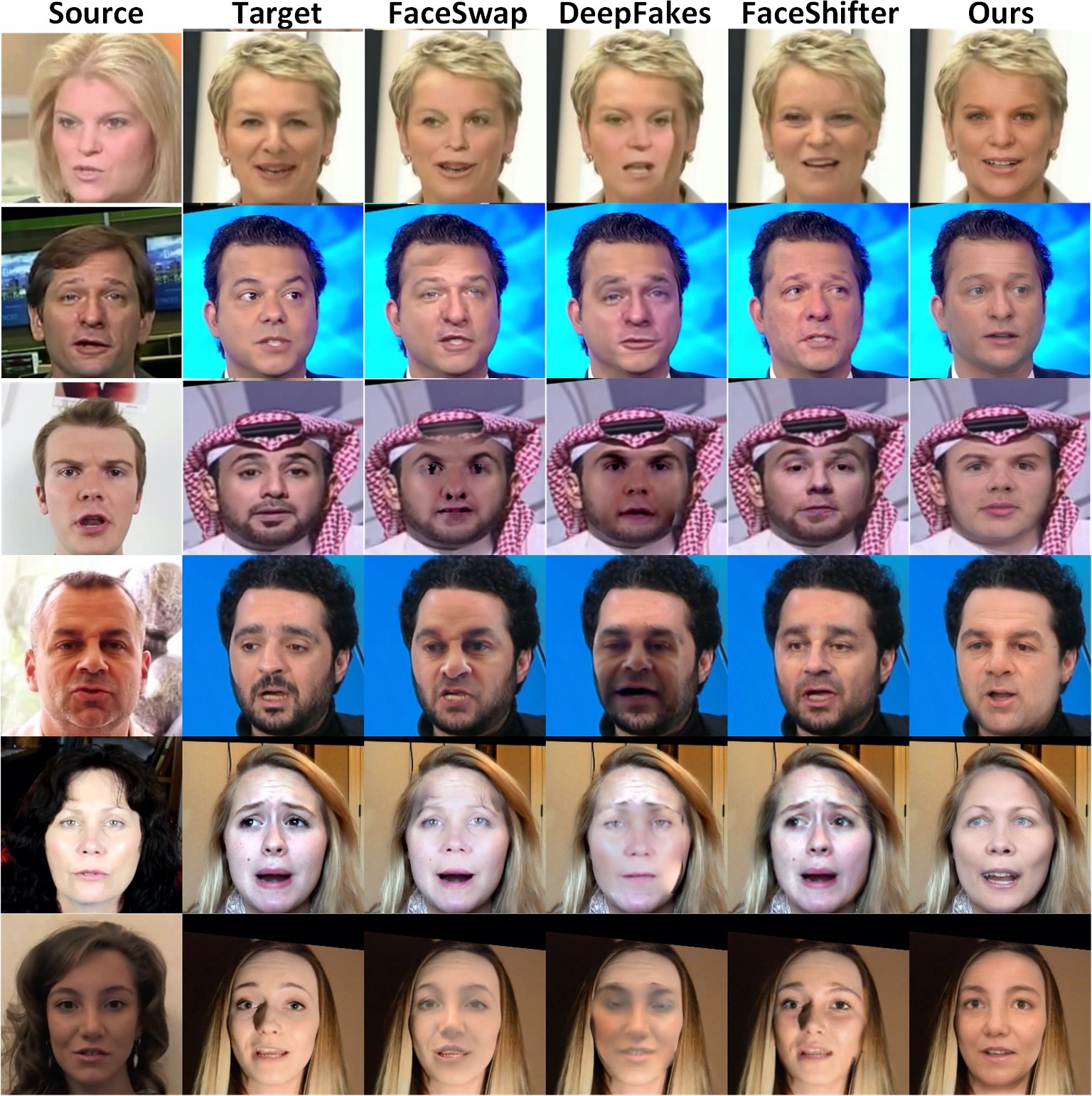}
    \caption{Qualitative comparison results of FaceSwap, DeepFakes, FaceShifter, and ours. FaceShifter and our method generate obviously better results than other methods. For FaceShifter, it generates wrong expressions in row 1 (fierce) and row 2 (fear), of which expressions are from source faces. Besides, FaceShifter keeps the beard of target faces in rows 3 and 4, which makes the swapped faces close to their target faces. In the last three rows, FaceShifter fails to swap faces. However, our method successfully preserves identity information from source images.}
    \label{fig:ffpp}
    \end{center}
\end{figure}

As shown in Fig.\ref{fig:ffpp}, FaceShifter and our method generate more visually pleasant results than other methods. For example, FaceSwap and DeepFakes suffer from blending inconsistency, distortions, and artifacts. For FaceShifter, the disentanglement of identity information from other attributes is sub-optimal because of its fixed identity encoder. In Fig.\ref{fig:ffpp}, FaceShifter generates unnatural expressions in the first and second rows, which seems to keep unnecessary expressions from the source images.
FaceShifter also fails to transfer identity information from source faces to target faces by incorrectly maintaining the beard from target faces in rows 3 and 4. Additionally as shown in rows 4, 5, and 6, FaceShifter tends to keep excessive attributes from target images, which makes the swapped faces similar to their target faces.

\vspace{0.2cm}
\noindent \textbf{Quantitative Comparison:}
In order to make a fair comparison with other methods quantitatively, we follow the experiment settings introduced in FaceShifter \cite{li2019faceshifter}. 

Firstly, ten frames per original video are evenly sampled and processed by MTCNN \cite{Zhang2016mtcnn}, resulting in 10,000 aligned faces. Then, aligned faces are manually checked in case of incorrect detections. After data cleaning, all corresponding frames in manipulated videos are extracted for testing. However, as FaceForensics++ is not designed for face recognition, some videos display repeated identities. For example, videos numbered $043$ and $343$ show Vladimir Putin, and videos of $179, 183$ and $826$ show the same person Barack Obama. Thus, we manually categorize all videos into $889$ identities. ID retrieval is measured as the top-1 matching rate of the swapped faces and their corresponding identities from source faces, serving to measure the identity preservation ability of different face swapping methods.
As for pose and expression errors, an open-sourced pose estimator \cite{Ruiz_2018_CVPR_Workshops} and a 3D facial model \cite{deng2019accurate} are used to extract pose and expression vectors. Then $\ell _2$ distances between swapped faces and the corresponding target faces are measured and recorded in Tab.\ref{tab:expffpp}.
\begin{table}[H]
    \small
    %\label{tab:expffpp1}
    \begin{center}
    \begin{tabular}{|l|c|c|c|}
    \hline
    Method          & ID retrieval $\uparrow$                & pose $\downarrow$                      & expression $\downarrow$\\
    \hline\hline
    DeepFakes \cite{Ondyari2019DeepFakes}      & 88.39                     & 4.64                     & 3.33 \\
    FaceSwap \cite{MarekKowalski2019FaceSwap}       & 72.69                     & 2.58                     & 2.89 \\
    Face2Face \cite{thies2016face2face}      & -                         & 2.68                     & \textbf{\textcolor{blue}{2.09}}\\
    Neural Textures \cite{thies2019deferred}& -                         & \textbf{\textcolor{red}{2.21}}    & \textbf{\textcolor{red}{1.64}} \\
    FaceShifter \cite{li2019faceshifter}    & \textbf{\textcolor{blue}{90.68}}   & \textbf{\textcolor{blue}{2.55}}   & 2.82 \\
    Ours            & \textbf{\textcolor{red}{90.83}}    & 2.64                     & 2.96 \\
    \hline
    \end{tabular}
    \end{center}
    %\label{tab:expffpp1}
    \caption{Quantitative comparison results on FaceForensics++. The best two results are shown in red and blue respectively. $\uparrow$ means higher is better, and $\downarrow$ means lower is better.}
    \label{tab:expffpp}
\end{table}

As DeepFakes, FaceSwap, and FaceShifter are face swapping methods, while Face2Face and Neural Textures are face reenactment methods, different evaluation criterions should be considered.  We report ID retrieval, pose error, and expression error for face swapping methods and neglect ID retrieval for face reenactment methods. 
As shown in Tab.\ref{tab:expffpp}, our method achieves the highest ID retrieval thanks to the hierarchical representation for faces. However, our method performs inferior to FaceShifter and reenactment methods in terms of pose and expression errors. Aside from face reenactment methods are mainly designed to control facial movements and expression deformations while neglecting to swap the identity information, two possible reasons hide behind. Firstly, our method is trained on only 500,000 images, which is much less than 2,700,000 images used to train FaceShifter. Besides, the training set for FaceShifter contains VGGFace \cite{parkhi2015deep}, which contains more pose and expression variations compared with CelebA-HQ and FFHQ. Secondly, StyleGAN2 is trained on FFHQ, which is proved to have data bias \cite{shen2020interfacegan}. Consequently, StyleGAN2 tends to generate smiling faces.
\begin{figure}[]
    \vspace{-0.2cm}
    \begin{center}
    \includegraphics[width=1.0\linewidth]{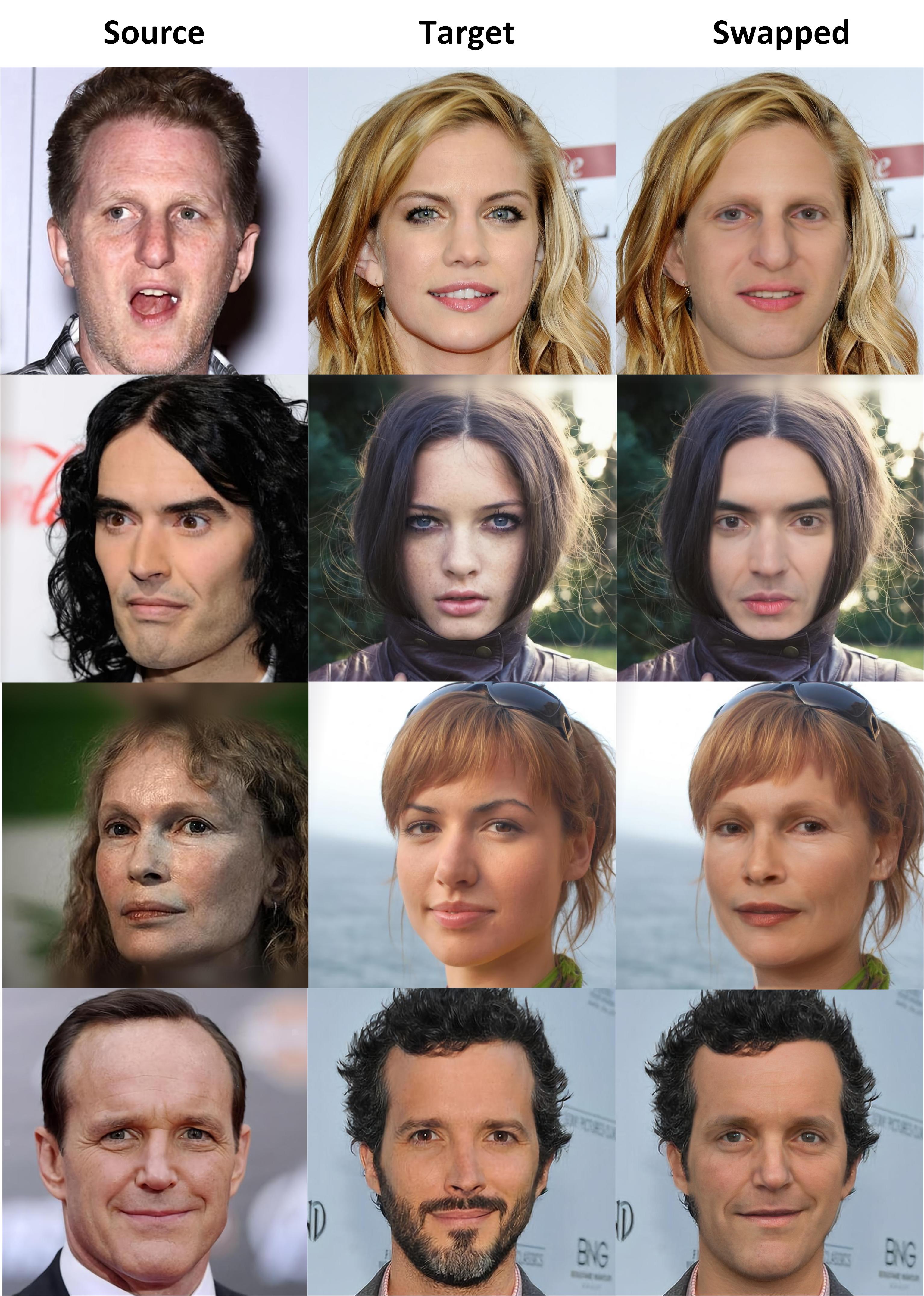}
    \caption{Face swapping results on CelabA-HQ. Images from right to left are source image $x_s$ which provides the identity, target image $x_t$ that offers the attributes, and the swapped face image $y_{s2t}$. All images are in $1024^2$.}
    \label{fig:celebahq}
    \end{center}
\end{figure}
\subsection{Experiments on CelebA-HQ} \label{subsec:celebahq}
\noindent \textbf{Qualitative Result: }
One superiority of our method is that it can achieve megapixel level face swapping.
As shown in Fig.\ref{fig:celebahq}, faces can be swapped across various expressions and poses. The swapped faces faithfully keep wrinkles, iris colors, eyebrow and nose shapes from source faces.
To the best of our knowledge, no other methods can swap faces at the resolution of $1024^2$ except for \cite{naruniec2020high}. However, \cite{naruniec2020high} needs to train different decoders for different identities, so it is not compared in this section.

\noindent \textbf{Quantitative Result:}
To quantify the capability of the proposed MegaFS on swapping megapixel face images, we randomly swapped 300,000 pairs of face images in CelebA-HQ for testing. For the reason that ID retrieval calculation between 30,000 original faces and 300,000 swapped faces requires \emph{Nine Billion} times of matching, we report cosine similarity of swapped faces and the corresponding source faces using cosface as ID similarity to release the computational burden. Also, both pose error and expression error are measured under the same settings as experimented in subsection \ref{subsec:expffpp}. In addition, Fr\'echet Inception Distance (FID) is reported to quantify the similarity of the 300,000 swapped face images to CelebA-HQ dataset. The results are summarized in Tab.\ref{tab:celebahq-p1} as the baseline for future research.
\begin{table}[]
    \small
    \begin{center}
    \begin{tabular}{|l|l|l|l|l|}
    \hline
    Method         & ID similarity $\uparrow$ & pose $\downarrow$ & expression $\downarrow$ & FID $\downarrow$\\
    \hline\hline
    Ours           & 0.5014             & 3.58    & 2.87   & 10.16         \\
    \hline
    \end{tabular}
    \end{center}
    \caption{Quantitative experimental results on CelebA-HQ. We report ID similarity, pose error, and expression error to demonstrate the megapixel level face swapping performance of the proposed MegaFS. FID is also reported as the similarity between the 300,000 swapped face images and CelebA-HQ dataset.}
    \label{tab:celebahq-p1}
\end{table}
%\begin{figure}[]
%%    \begin{center}
%    \includegraphics[width=1.0\linewidth]{images/wp-wpp-1.png}
%    \caption{The comparison between latent space $\mathcal{W}^+$ and $\mathcal{W}^{++}$: easy case.}
%    \label{fig:wp-wpp-1}
%    \end{center}
%\end{figure}

\subsection{Ablation Study} \label{subsec:ablation}
In this section, we conduct ablation experiments on CelebA-HQ to evaluate the effectiveness of the key components in the proposed MegaFS.
%\vspace{-0.9cm}
\subsubsection{The Choise of Latent Space}
In this part, we will verify the superiority of the extended latent space $\mathcal{W}^{++}$ over $\mathcal{W}^{+}$. We trained another neural network, which has the same network structure as HieRFE, to project facial images into latent space $\mathcal{W}^+$.%Hyper-parameters are same as in subsection \ref{subsec:did}. %Training data consists of FFHQ and the 200K auxiliary data. Two models are evaluated on CelebA-HQ.
\begin{table}[H]
    \begin{center}
    \begin{tabular}{|l|c|c|c|}
    \hline
    Latent Space &  LPIPS $\downarrow$ & MSE $\downarrow$ & failure rate $\downarrow$ \\
    \hline\hline
    $\mathcal{W}^+$        & 0.2486             & 0.0672    & 1.28\%  \\
    $\mathcal{W}^{++}$     & \textbf{\textcolor{red}{0.2335}}  & \textbf{\textcolor{red}{0.0563}}  & \textbf{\textcolor{red}{0.65\%}}    \\
    \hline
\end{tabular}
\end{center}
\caption{Quantitative comparison results of latent space $\mathcal{W}^+$ and $\mathcal{W}^{++}$ using GAN Inversion metrics. LPIPS $\ell _2$ distance and image level MSE are measured to quantify the information preservation capabilities of $\mathcal{W}^+$ and $\mathcal{W}^{++}$. The robustness is indicated by the failure rate of facial reconstruction. For reported metrics, HieRFE outperforms its counterpart trained on $\mathcal{W}^+$.}
\label{tab:wp-wpp-p2}
\end{table}

For illustrating the information preservation ability, two widely used metrics in GAN Inversion, LPIPS $\ell _2$ distance and image level MSE, are reported in Tab.\ref{tab:wp-wpp-p2}. Besides, the percentage of unsuccessful reconstructions is defined as the failure rate to quantify the robustness of two inversion models. From the reported results, we can conclude that HieRFE outperforms its counterpart trained on $\mathcal{W}^+$ for better information preservation ability as well as the robustness.
As to the controllability, we use the same evaluation criterions in subsection \ref{subsec:celebahq} to evaluate different latent space.
The quantitative results are shown in Tab.\ref{tab:wp-wpp-p1}, suggesting that $\mathcal{W}^{++}$ is better than $\mathcal{W}^+$ in terms of ID similarity, pose and expression preservation ability.

%In addition, two inversion models achieve similar FID, SSIM, and MS-SSIM as reported in Tab.\ref{tab:wp-wpp-p3}, measuring the similarity between the CelebA-HQ dataset and the reconstructed images.

%Finally, we report FID, SSIM, and MS-SSIM in Tab.\ref{tab:wp-wpp-p3} to measure the similarity between the CelebA-HQ dataset and the reconstructed images with backgrounds excluded. It seems both models for $\mathcal{W}^+$ and $\mathcal{W}^{++}$ projection perform well. However, please note that these metrics only measured on the successfully reconstructed images, which means the inversion model for $\mathcal{W}^{++}$ achieves a similar result even considering more difficult samples.
\begin{figure}[]
    %\vspace{-0.2cm}
    \begin{center}
    \includegraphics[width=1.0\linewidth]{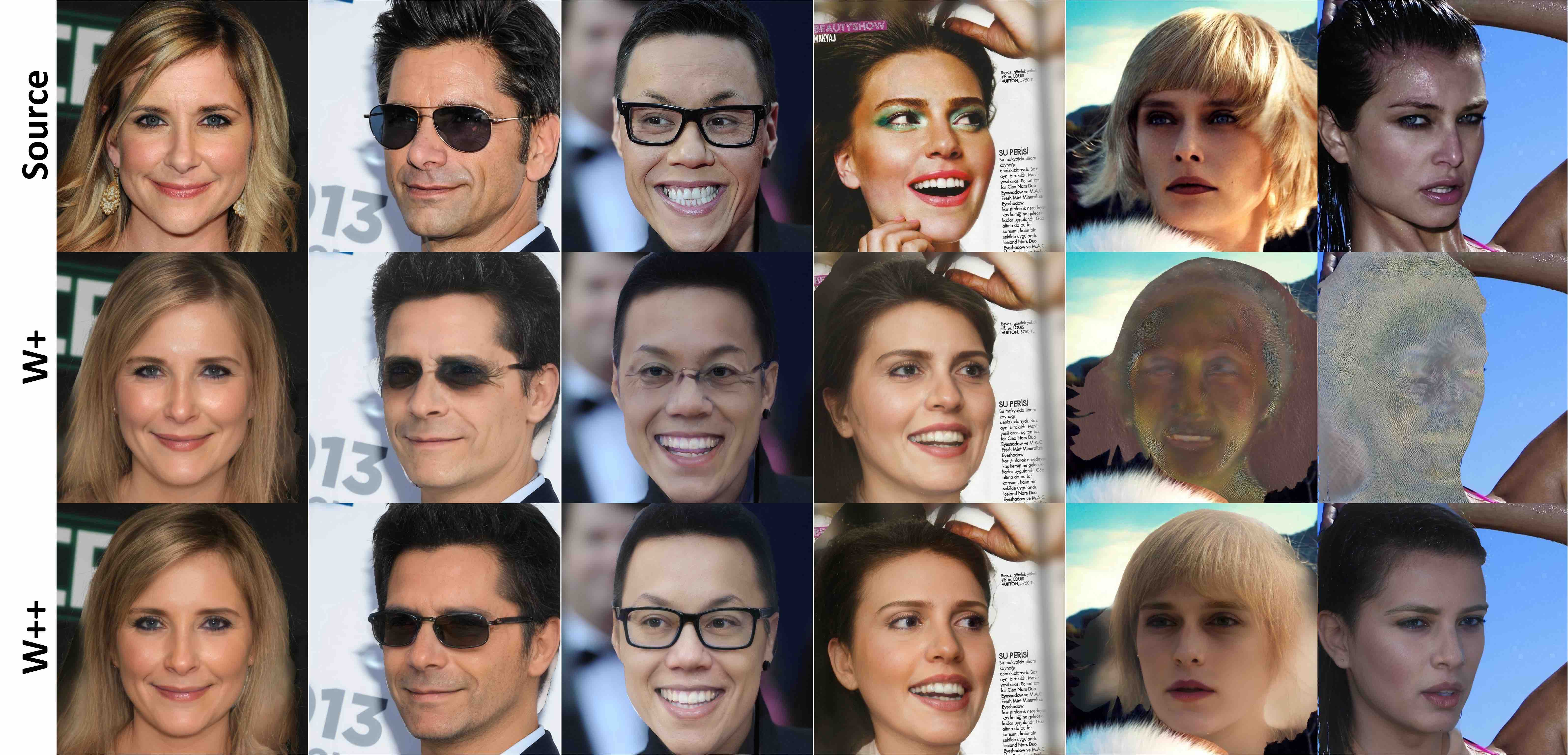}
    \caption{Qualitative comparison results of reconstructed images from latent space $\mathcal{W}^+$ and $\mathcal{W}^{++}$. From top to bottom: source images, reconstructed images from $\mathcal{W}^+$ and $\mathcal{W}^{++}$. HieRFE and its counterpart perform well in easy cases (the first column), but the latter fails to recast sunglasses, glasses, eye gazes, and faces under complex lighting conditions (from the second to the last columns).}
    \label{fig:wp-wpp-2}
    \end{center}
\end{figure}
\begin{table}[]
    %\small
    \begin{center}
    \begin{tabular}{|l|c|c|c|}
    \hline
    Latent Space & ID similarity $\uparrow$ & pose $\downarrow$ & expression $\downarrow$ \\
    \hline\hline
    $\mathcal{W}^+$        & 0.5438        & 4.0640    & 1.7467           \\
    $\mathcal{W}^{++}$     & \textbf{\textcolor{red}{0.5816}}  & \textbf{\textcolor{red}{3.8179}}  & \textbf{\textcolor{red}{1.6489}}          \\
    \hline
\end{tabular}
\end{center}
\caption{Quantitative comparison results of latent space $\mathcal{W}^+$ and $\mathcal{W}^{++}$ using face swapping metrics. HieRFE trained on $\mathcal{W}^{++}$ beats its counterpart trained on $\mathcal{W}^+$ in terms of ID similarity, pose error, and expression error.}
\label{tab:wp-wpp-p1}
\end{table}
%\begin{table}[]
%    \begin{center}
%    \begin{tabular}{|l|c|c|c|}
%    \hline
%    Latent Space & FID $\downarrow$ & SSIM $\uparrow$ & MS-SSIM $\uparrow$ \\
%    \hline\hline
%    $\mathcal{W}^+$        & \textcolor{red}{10.38}             & 0.72    & 0.70      \\
%    $\mathcal{W}^{++}$     & 10.42             & \textcolor{red}{0.73}    & \textcolor{red}{0.71}      \\
%    \hline
%\end{tabular}
%\caption{The similarity between the CelebA-HQ data and corresponding reconstructed images from $\mathcal{W}^+$ and $\mathcal{W}^{++}$, in terms of FID, SSIM, and MS-SSIM. Images reconstructed from both latent space have very close similarities compared to CelebA-HQ. However, the inversion model trained on $\mathcal{W}^{++}$ can reconstruct more difficult samples.}
%\label{tab:wp-wpp-p3}
%\end{center}
%\end{table}
The qualitative results of two inversion models are displayed in Fig.\ref{fig:wp-wpp-2}. HieRFE and its counterpart can reconstruct easy cases well. However, the latter fails to recast sunglasses, eyeglasses, eye gazes, and faces under complex lighting conditions. Thus, the latent space $\mathcal{W}^{++}$ is verified to be better than $\mathcal{W}^+$ for both face reconstruction and face swapping tasks in all terms of information preservation ability, robustness, and controllability.
%in the GAN Inversion task and is more suitable for face swapping task for its better controllability.

%\subsubsection{The Necessity of Landmark Supervision}
%For landmark loss $\mathcal{L}_{ldm}$, it supervises FTM to focus on details that can be easily neglected such as eye states, gaze directions, and shapes of mouth region. The qualitative results are shown in Fig.\ref{fig:ablationLldm}.
%\small
%\begin{figure}[h]
%    \begin{center}
%    \includegraphics[width=0.9\linewidth]{images/aaa.jpg}
%    \caption{Without $\mathcal{L}_{ldm}$, FTM cannot preserve expression details of the target on swapped faces, inlucding eye states (open or closed), gaze directions, and mouth shapes.}
%    \label{fig:ablationLldm}
%\end{center}
%\end{figure}

%\vspace{-1cm}
%\subsubsection{The Discussion of $C,$ $L^{low}$ and $L^{high}$}
%To envisage the functionality of $C,$ $L^{low}$ and $L^{high}$, we make use of $[C_t, L_t^{high}, \bm{L_s^{high}}]$ instead of $[C_t, L_t^{high}, \bm{L_{s2t}}]$ for generation through StyleGAN2. The reuslts are shown in the third column of Fig.\ref{fig:ftmvs}, marked as Latent Code Replacement (LCR). 

%Evidently, LCR can swap faces while ignoring target attributes such as skin color and eye state. Thus, we can safely conclude that identity information, to a large extent, are encoded in $L^{high}$. Thus, $C_s$ and $L_s^{low}$ are discarded in the proposed pipeline.

\subsubsection{The Design of Latent Code Manipulator}
\begin{figure}[]
    %\vspace{-0.2cm}
    \begin{center}
    \includegraphics[width=1.0\linewidth]{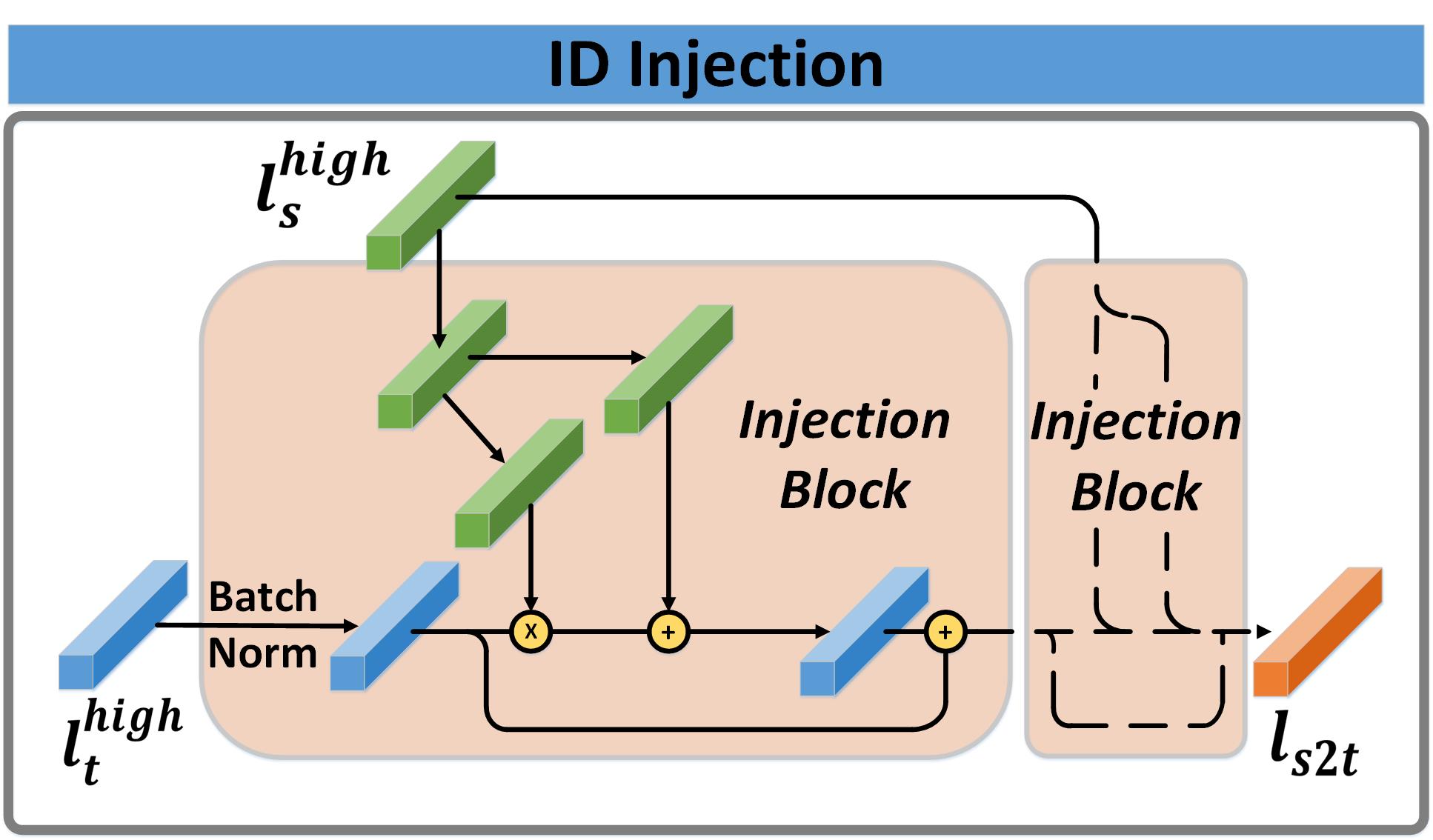}
    \caption{The design of ID Injection follows the SPADE ResBlk \cite{park2019semantic}, with convolutional layers for 2D input subsititued by linear layers, indicated by black arrows, for vectors.}
    \label{fig:idInject}
    \end{center}
\end{figure}

As StyleGAN2 has a layer-wise representation \cite{collins2020editing, karras2019style, karras2020analyzing}, it is heuristically feasible to manipulate latent codes by any network that operates on vectors. However, we argue that the design of the latent code manipulator needs to consider the applicability on vectorized information exchanging. 
To this end, we make use of $[C_t, L_t^{high}, \bm{L_s^{high}}]$ instead of $[C_t, L_t^{high}, \bm{L_{s2t}}]$ for generation, named as Latent Code Replacement (LCR), to envisage the functionality of $C,$ $L^{low}$ and $L^{high}$. 
Afterwards, we follow the previous method \cite{park2019semantic} to inject identity information into latent codes. This design is detailed in Fig.\ref{fig:idInject}, namely ID Injection. Both designs are compared to the proposed FTM.

\begin{figure}[]
    %\vspace{-0.35cm}
    \begin{center}
    \includegraphics[width=1.0\linewidth]{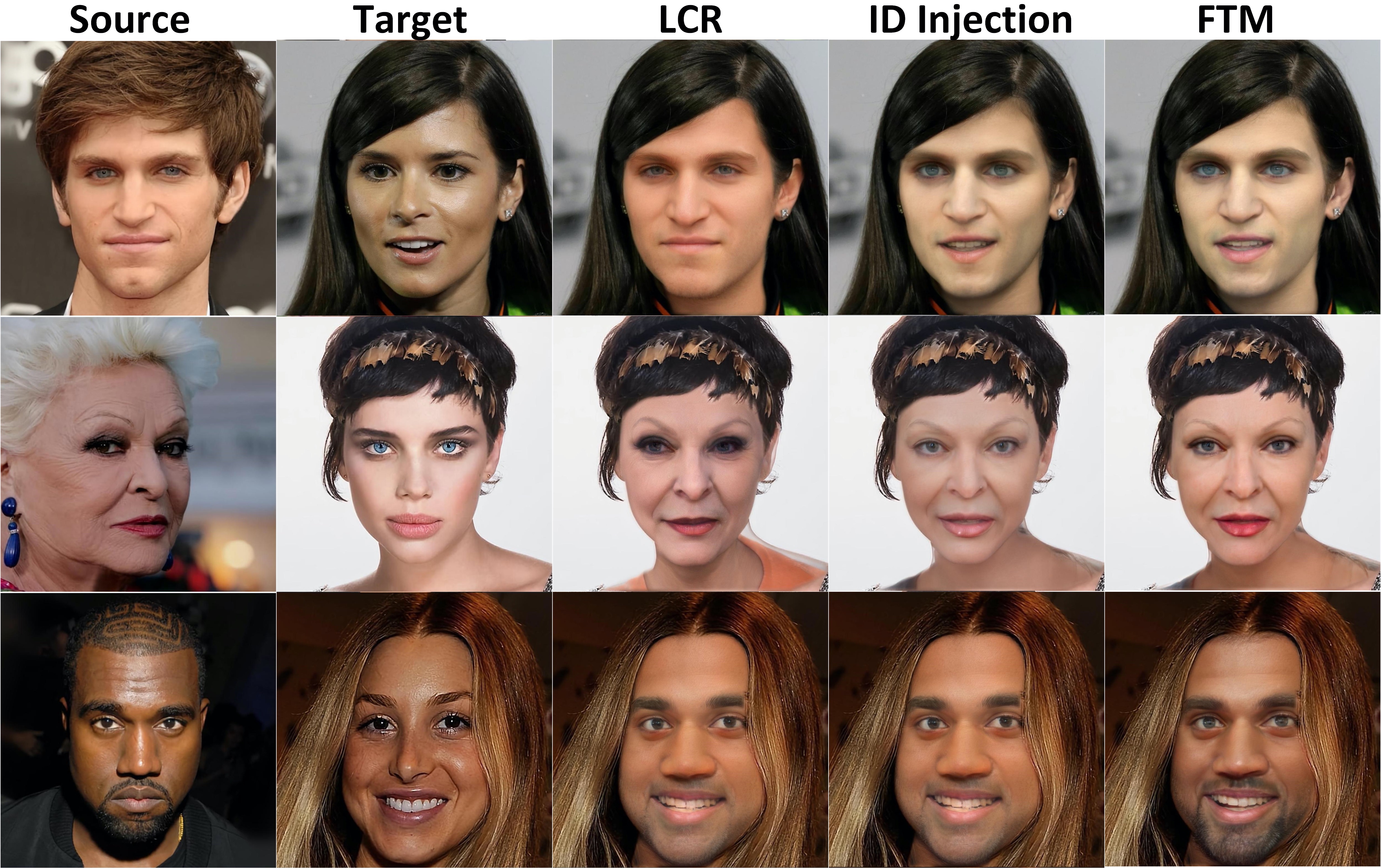}
    \caption{Qualitative comparison results of different latent code manipulation methods: LCR, ID Injection, and FTM. LCR keeps skin color and eye state from the source image $x_s$ as shown in the first two rows. For ID Injection, attributes are dominated by the target image $x_t$. For example, the red lip in row 2 and beard in the last row from source images are neglected. FTM achieves the best balance among the three latent code manipulation methods.}
    \label{fig:ftmvs}
    \end{center}
\end{figure}

For a fair comparison, other two sets of 300,000 swapped face images are generated by adopting LCR and ID Injection respectively. The quantitative results are summarized in Tab.\ref{tab:latentcontrol} and the qualitative comparison is shown in Fig.\ref{fig:ftmvs}.
LCR achieves the best FID since it keeps excessive semantic information from source images. However, this is not favorable for face swapping since information from $L_t^{high}$ is lost. As shown in the third column of Fig.\ref{fig:ftmvs}, LCR can swap faces while ignoring target attributes such as skin color and eye state. Thus, we can safely conclude that identity information, to a large extent, is encoded in $L^{high}$. Thus, $C_s$ and $L_s^{low}$ are discarded in the proposed pipeline.

Based on this observation, ID Injection and FTM are proposed to process only on $L^{high}$. For ID Injection, it has the lowest expression error as it reserves topology information from target images at the cost of other semantic parts from source images, such as identity information stated in Tab.\ref{tab:latentcontrol}, and facial details as shown in the fourth column of Fig.\ref{fig:ftmvs}. Among them, FTM achieves the highest ID similarity level, lowest pose error, and makes a decent balance in terms of expression error, FID, and visual pleasantness. Thus, the proposed FTM shows to be better than the other two latent code manipulators for face swapping.
%\begin{figure}[]
%    \begin{center}
%        \includegraphics[width=0.9\linewidth]{images/LM.png}
        %\subfigure[Latent Code Replacement]{
        %    \begin{minipage}{0.45\linewidth}
        %        \begin{center}
        %        \includegraphics[width=1.05\linewidth]{images/lcr.png}
        %    \end{center}
        %    \end{minipage}
        %}
        %\subfigure[ID Injection]{
        %    \begin{minipage}{0.49\linewidth}
        %        \begin{center}
        %        \includegraphics[width=1.05\linewidth]{images/ij.png}
        %        \end{center}
        %    \end{minipage}
        %}
%    \caption{(a) the $L_t^{high}$ is replaced by $L_s^{high}$; (b) the design follows the SPADE ResBlk \cite{park2019semantic}, with convolutional layers for 2D input subsititued by linear layers for vectors, indicated by black arrows.}
%    \label{fig:lm}
%%    \end{center}
%\end{figure}
\begin{table}[]
    %\vspace{-0.5cm}
    \small
    \begin{center}
    \begin{tabular}{|l|c|c|c|c|}
    \hline
    Latent Control & ID similarity $\uparrow$ & pose $\downarrow$ & exp $\downarrow$ & FID $\downarrow$ \\
    \hline\hline
    LCR  & 0.3997             & 5.04    & 3.43          & \textbf{\textcolor{red}{9.64}}          \\
    ID Injection & 0.4447             & 3.67    & \textbf{\textcolor{red}{2.82}}          & 10.32        \\
    FTM (Ours)     & \textbf{\textcolor{red}{0.5014}} & \textbf{\textcolor{red}{3.58}}    & 2.87          & 10.16       \\
    \hline
    \end{tabular}
    \end{center}
    \caption{Quantitative comparison results of different latent code manipulation methods (``exp'' represents expression error). FTM achieves the best ID similarity and pose preservation results and makes a decent balance among expression error and FID.}
    \label{tab:latentcontrol}
\end{table}
%\newpage
\section{Conclusion}
In this paper, we have analyzed three unsettled key issues in previous works for high resolution face swapping and proposed a general face swapping pipeline named MegaFS to resolve these difficulties in a three-stage procedure. HieRFE in the first stage projects faces into hierarchical representaions in an extended latent space $\mathcal{W}^{++}$ for complete facial information deposit. FTM in the second stage transfers the identity from a source image to the target by a non-linear trajectory without explicit feature disentanglement. Finally, StyleGAN2 is used to synthesize the swapped face and avoid unstable adversarial training. The modular design of MegaFS requires little GPU memory with a negligible performance cost and it performs comparatively when compared to other state-of-the-art face swapping methods at the resolution of $256^2$. Besides, to the best of our knowledge, MegaFS is the first method that can conduct one shot face swapping on megapixels. Finally, based on MegaFS, the first megapixel level face swapping database is built and released to the public for future research of forgery detection and face swapping.

%-------------------------------------------------------------------------
\section{Acknowledgements}
This work was supported in part by the National Key R$\&$D Program of China under Grant 2020AAA0140002, in part by the Science and Technology Development Fund of Macau SAR under grant 0015/2019/AKP, in part by the Natural Science Foundation of China under Grant 62076240, Grant 61721004, Grant U1836217, and in part by the Macao Youth Scholars Program under Grant AM201913.

\newpage
{\small
\bibliographystyle{ieee_fullname}
\bibliography{egbib}
}

\end{document}